# Diseño y Desarrollo de Prototipos Robóticos Para Competencias de Fútbol Utilizando Motores Dynamixel

Moraes Pablo, pablo.moraes@estudiantes.utec.edu.uy[1]
Sodre Hiago, hiago.sodre@estudiantes.utec.edu.uyl[2]
Rodríguez Monica, monica.rodriguez@utec.edu.uy[3]
Kelbouscas, Andre, andre.dasilva@utec.edu.uy[4]
Jean Schuster, jean.schuster@utec.edu.uy[5]
Crsitiano Schuster, crsitiano.schuster@utec.edu.uy[6]
Grando Ricardo, ricardo.bedin@utec.edu.uy[7]
[1]Universidad Tecnológica del Uruguay
[2]Universidad Tecnológica del Uruguay
[3]Universidad Tecnológica del Uruguay
[4]Universidad Tecnológica del Uruguay
[5]Universidad Tecnológica del Uruguay
[6]Universidad Tecnológica del Uruguay
[7]Universidad Tecnológica del Uruguay

***Abstract:*** *This article describes the design and development of robotic prototypes for robotic soccer competitions using Dynamixel motors. Although the prototypes are not aimed at world-class competitions, they represent a significant step in the development of sports robots. Model XL430-W250 Dynamixel motors were chosen and electronic circuits were implemented using control boards such as OpenCR and Raspberry Pi 3. A crucial component was introduced: a step-up board that charges a capacitor to create a powerful kick to the ball via an electromagnet controlled by Arduino Nano. The programming and coordination of the prototypes was carried out using the ROS environment (Robot Operating System), which allows effective integration of movements and communication. Although the prototypes were not optimized for global competition, they underwent extensive testing, evaluating their speed and maneuverability, as well as soccer tactics in the GRSim simulator. These prototypes contribute to the further development of sports robotics and illustrate the research potential in this exciting area.*

***Resumen:*** *Este artículo describe el diseño y desarrollo de prototipos robóticos para competiciones de fútbol robótico utilizando motores Dynamixel. Aunque los prototipos no están destinados a competiciones de talla mundial, representan un paso significativo en el desarrollo de robots deportivos. Se eligieron motores Dynamixel modelo XL430-W250 y se implementan circuitos electrónicos utilizando tableros de control como OpenCR y Raspberry Pi 3. Se introdujo un componente crucial: una placa elevadora que carga un capacitor para crear una poderosa patada a la pelota a través de un electroimán. controlado por Arduino Nano. La programación y coordinación de los prototipos se realizó utilizando el entorno ROS (Robot Operating System), que permite una integración efectiva de movimientos y comunicación. Aunque los prototipos no fueron optimizados para la competencia global, se sometieron a pruebas exhaustivas, evaluando su velocidad y maniobrabilidad, así como las tácticas de fútbol en el simulador GRSim. Estos prototipos contribuyen al desarrollo de la robótica deportiva e ilustran el potencial de investigación en este apasionante campo.*

Palabras clave: *Prototipos de robots, Motores Dynamixel, Competencias de fútbol de robots*

# 1 - INTRODUCCIÓN

La robótica y la inteligencia artificial han impulsado transformaciones en diversos sectores, generando avances notables en la interacción con la tecnología y la resolución de desafíos complejos. En este contexto de innovación, este artículo documenta el diseño y desarrollo de un prototipo de robots diseñados para competiciones de fútbol de robots, con un enfoque especial en la integración de motores Dynamixel. Nuestro enfoque se ha centrado en la así como la agilidad, robustez y el rendimiento en el entorno de juego.

En este artículo, se detalla la implementación de circuitos electrónicos que incorporan placas de control como OpenCL 1.0 y Raspberry Pi 3. Además, introducimos una placa step-up que desempeña el papel fundamental de cargar un capacitor para generar una patada a la pelota mediante un sistema de electromagnético controlado por un Arduino Nano. El control de los prototipos se llevó a cabo mediante una programación cuidadosa en el entorno de desarrollo ROS (Robot Operating System), que proporciona una estructura modular y flexible para el desarrollo de aplicaciones robóticas. Utilizamos ROS para programar y coordinar los movimientos de los motores Dynamixel, así como para establecer la comunicación entre las diferentes placas y dispositivos.

A pesar de que estos prototipos no han sido optimizados para competencias de categoría mundial, los sometimos a pruebas exhaustivas en condiciones reales. Evaluamos su velocidad, maniobrabilidad y respuestas ante situaciones específicas de juego. Además, empleamos el simulador GRSim para concebir y evaluar tácticas de fútbol a medida. Si bien estos prototipos no representan el cenit de la competitividad a nivel global, sentaron las bases para futuras mejoras y enriquecieron la comunidad científica y robótica con su contribución al diseño y desarrollo de robots deportivos.

Este trabajo no sólo representa un avance en el campo de la robótica deportiva, sino que también demuestra el potencial de investigación en este campo. Al refinar y optimizar cuidadosamente nuestros prototipos, allanamos el camino hacia la próxima generación de robots de fútbol de alto rendimiento. La metodología de robot propuesta fue desarrollada durante el año de 2022 en un proyecto con la embajada de los Estados Unidos en Uruguay y fue validado a través de una competencia local en abril de 2023. Para una transparencia total y para aquellos que quieran más detalles, todo el proceso ha sido documentado y compartido en GitHub [enlace al repositorio de GitHub].

# 2 - METODOLOGÍA

La metodología propuesta para este artículo está dividida de manera progresiva y cronológica de trabajo realizado por el mismo, se tomó una secuencia de diseño mecánico, presentación de los motores dynamixel, electrónica y circuitos de los robots, programación y control y ambiente final de simulación.

2.1. Diseño Mecánico

El diseño mecánico de los robots fue un proceso integral en el que se prioriza la agilidad, la durabilidad y el rendimiento en el campo de juego. Cada robot presenta una estructura compacta y aerodinámica para optimizar la maniobrabilidad, como su modelo de diseño en formato CAD que se muestra en la Figura 1. La distribución de los componentes se diseñó de manera estratégica, con un enfoque en el equilibrio y la estabilidad. Se emplearon sistemas de suspensión en las ruedas para garantizar la tracción en terrenos diversos. (Ver Figura 3.1 para un diagrama detallado del diseño).

En el contexto del diseño mecánico de los prototipos de robots para competiciones de fútbol de robots, se implementó una estrategia específica para optimizar la velocidad y agilidad de los movimientos. Para lograr un aumento en la velocidad de propulsión, se recurrió a la integración de engranajes de mayor tamaño conectados a los motores Dynamixel XL430-W250. Esta elección de diseño se fundamenta en el principio mecánico de multiplicación de velocidad a través de la relación de engranajes.

Al utilizar engranajes de mayor diámetro, se logra un efecto de multiplicación en la velocidad de rotación entregada a las ruedas del robot. Este enfoque mecánico permite que los motores operen a un régimen óptimo mientras proporcionan una salida de velocidad superior en comparación con los engranajes estándar. La ventaja resultante radica en la capacidad de los prototipos para moverse con mayor velocidad en el campo de juego, lo que es esencial para ejecutar maniobras ágiles y responder de manera eficaz a las demandas de juego.

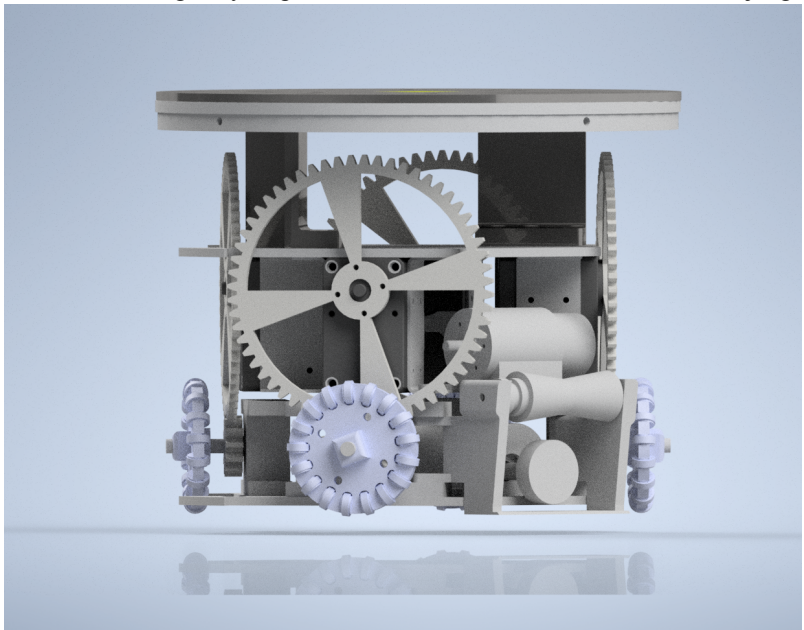

Figura 1. Modelo CAD del robot SSL.

2.2. Motores Dynamixel

La elección de los motores Dynamixel XL 430-W 250, que se ilustran en la Figura 2, permitió una mayor coherencia con el hardware y las tecnologías utilizadas en otros proyectos de la comunidad robótica, facilitando la integración y el intercambio de conocimientos. La compatibilidad con los TurtleBots 3 y otras plataformas robustas respalda la confiabilidad y la estandarización en el desarrollo de sistemas robóticos, así como la ampliación de los beneficios derivados de la experiencia colectiva en el uso de estos motores en diversos contextos.

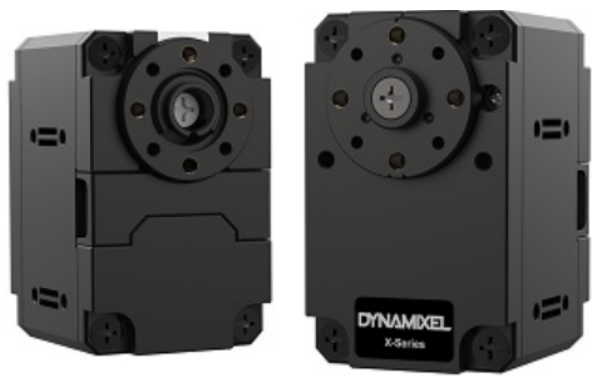

Figura 2. Imagen ilustrativa de motores Dynamixel X-series.

La selección de estos motores Dynamixel fortaleció la cohesión entre los componentes electrónicos y mecánicos, y su versatilidad demostrada en aplicaciones previas subraya su idoneidad para potenciar la agilidad y la precisión de los prototipos de robots de fútbol en nuestro proyecto. Se optó por estos motores debido a su probada eficacia en aplicaciones de robótica móvil, Estos motores son conocidos por su precisión en la posición, control de velocidad y alto torque, lo que se alinea con los requisitos de nuestra aplicación en la robótica futbolística. Su capacidad para adaptarse a cambios rápidos en la dirección y velocidad, así como su respuesta precisa, resultaron ser importantes para lograr movimientos fluidos y la implementación efectiva de estrategias en el entorno de competición.

2.3. Electrónica y Circuitos

Cada robot cuenta con un conjunto de circuitos personalizados diseñados para manejar la comunicación, el control y la energía, que se ven en la Figura 3. La placa de control principal es la OpenCR, que incorpora un microcontrolador STM32 para recibir los datos procesados por la raspberry pi, ya que esta tiene dentro el software de control que filtra los datos recibidos de la simulación y envía los valores de velocidad via serial hacia la placa OpenCR, luego ella se encarga de mover los motores.

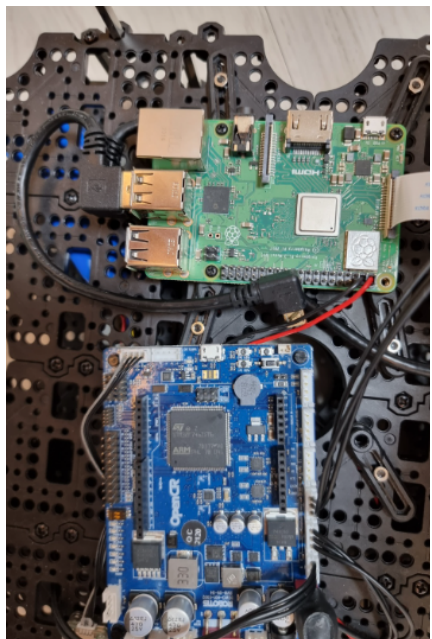

Figura 3. Placas utilizadas en el robot.

Los motores Dynamixel están interconectados a través de un bus RS-485, permitiendo la comunicación simultánea y la sincronización precisa de los movimientos. Para el procesamiento de alto nivel y la toma de decisiones, se utilizó una Raspberry Pi 3. Además, se implementó un step-up de 12V a 190V para el mecanismo de pateo. Este step-up carga un capacitor y luego se utiliza el Arduino Nano para activar un relé que descarga el capacitor en un sistema de electroimán, generando una patada en la pelota. (Ver Figura 3.3 Placas principales, Raspberry pi 3 y OpenCR 1.0).

2.4. Programación y Control

El control de los motores Dynamixel y la coordinación entre las diferentes placas se lograron mediante una programación cuidadosa, que se muestra una parte en la Figura 4 a continuación. Se utilizó el entorno de programación Arduino para desarrollar el software de control en la OpenCR. La Raspberry Pi 3 fue utilizada para la visión por computadora y el procesamiento de datos de alto nivel.

```python
        else:
            print("Dynamixel has been successfully connected")

        # Write goal position
        if (MY_DXL == 'XL320'): # XL320 uses 2 byte Position Data, Check the size of data in your DYNAMIXEL's control table
            dxl_comm_result, dxl_error = packetHandler.write2ByteTxRx(portHandler, DXL_ID, ADDR_GOAL_POSITION, dxl_goal_position[index])
        else:
            dxl_comm_result, dxl_error = packetHandler.write4ByteTxRx(portHandler, DXL_ID, ADDR_GOAL_POSITION, dxl_goal_position[index])
        if dxl_comm_result != COMM_SUCCESS:
            print("%s" % packetHandler.getTxRxResult(dxl_comm_result))
        elif dxl_error != 0:
            print("%s" % packetHandler.getRxPacketError(dxl_error))

def read_pwm_motors(DXL_ID):
    # Read present position
    if (MY_DXL == 'XL320'): # XL320 uses 2 byte Position Data, Check the size of data in your DYNAMIXEL's control table
        dxl_present_position, dxl_comm_result, dxl_error = packetHandler.read2ByteTxRx(portHandler, DXL_ID, ADDR_PRESENT_POSITION)
    else:
        dxl_present_position, dxl_comm_result, dxl_error = packetHandler.read4ByteTxRx(portHandler, DXL_ID, ADDR_PRESENT_POSITION)
    if dxl_comm_result != COMM_SUCCESS:
        print("%s" % packetHandler.getTxRxResult(dxl_comm_result))
    elif dxl_error != 0:
        print("%s" % packetHandler.getRxPacketError(dxl_error))

    print("[ID:%03d] GoalPos:%03d  PresPos:%03d" % (DXL_ID, dxl_goal_position[index], dxl_present_position))

    return dxl_present_position

def pwm_motor_callback(data):
    apply_pwm(DXL_ID_1, data.data[0])
    apply_pwm(DXL_ID_2, data.data[1])
    apply_pwm(DXL_ID_3, data.data[2])
    apply_pwm(DXL_ID_4, data.data[3])

def pwm_motor_2_callback(data):
    apply_pwm(DXL_ID_2, data.data[0])

def pwm_motor_3_callback(data):
    apply_pwm(DXL_ID_3, data.data[0])

def pwm_motor_4_callback(data):
    apply_pwm(DXL_ID_4, data.data[0])

if __name__=="__main__":
    rospy.init_node("dynamixel_interface_node", anonymous=False)

    index = 0
    dxl_goal_position = [1500, DXL_MAXIMUM_POSITION_VALUE]          # Goal position
```

Figura 4. Parte del código en lenguaje Python con ROS para los robots.

Se estableció una comunicación entre la OpenCR y la Raspberry Pi 3 para lograr una sincronización fluida y un control preciso de los movimientos. (Ver Figura 3.4 para una captura de pantalla de la interfaz de programación).

## 3 - RESULTADOS Y DISCUSIÓN

El rendimiento de los prototipos se evaluó en un entorno de prueba realista, simulando condiciones de juego. Se llevaron a cabo pruebas de velocidad, maniobrabilidad y respuesta a situaciones específicas de la competición, visibles en la Figura 6. Se midieron las métricas de tiempo y precisión para cuantificar el rendimiento de los robots en diferentes escenarios.

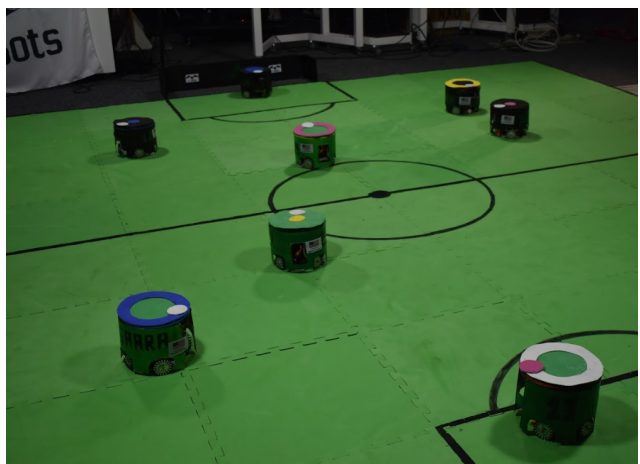

Figura 6. Juego de Fútbol con robots desarrollados.

Las conclusiones extraídas de los resultados y discusiones presentados anteriormente proporcionan una base sólida para evaluar el éxito y el impacto de este proyecto para desarrollar prototipos de robots para competiciones de fútbol robótico. En esta sección, resumimos los logros alcanzados, reflexionamos sobre las implicaciones de estos logros y esbozamos posibles direcciones futuras para ampliar aún más la investigación en esta área.

Muestra cómo la implementación de los prototipos con motores Dynamixel y el enfoque en la táctica del fútbol aumentaron la comprensión de las capacidades y limitaciones de los robots en un contexto competitivo. Además, el evento URUCUP, donde los robots fueron usados para validación, destaca la importancia de la interacción y el intercambio de conocimientos en la comunidad robótica global.

En cuanto a las direcciones futuras, consideramos el potencial de mejorar aún más el rendimiento de los robots mediante la optimización continua de las tácticas y la coordinación en el campo. La integración de nuevas tecnologías como la inteligencia artificial y el aprendizaje automático podría ser una forma prometedora de lograr mayores niveles de autonomía y adaptabilidad en los robots. Además, el desarrollo de colaboraciones y competiciones regionales podría enriquecer aún más la experiencia de los equipos y promover el intercambio de ideas.

En última instancia, este proyecto sienta las bases para un futuro en la robótica deportiva y sus aplicaciones. A través de la exploración constante y la colaboración global, estamos motivados para evolucionar, inspirar y superar los límites de la tecnología robótica en el deporte.

## 4 - CONCLUSIONES

En este trabajo fue presentado el diseño de prototipos de robots para competiciones de fútbol de robots. Se utilizaron motores Dynamixel XL 430-W 250, esos componentes mecánicos avanzados marcaron una diferencia notable en el desarrollo de los prototipos. Los sistemas electrónicos sofisticados también fueron parte del proceso. También fueron utilizadas estrategias de programación cuidadosamente conceptualizadas. Estos elementos combinados dieron como resultado prototipos ágiles. Son capaces de ejecutar tácticas de fútbol tanto en escenarios simulados como reales.

Los aspectos críticos del éxito fueron la implementación de tácticas específicas. Un factor fue la coordinación eficiente de los movimientos de los robots. Fueron adoptadas tecnologías como (ROS) Sistema Operativo de Robots. También utilizamos una Raspberry Pi 3 junto con la placa OpenCR. Estos desempeñaron un papel clave en la ejecución de los movimientos. También ayudaron a implementar estrategias en entornos dinámicos y altamente competitivos.

Los resultados de las evaluaciones realizadas con simulaciones en diferentes momentos dieron sus frutos. Se obtuvieron conocimientos sobre las capacidades y las limitaciones de nuestros prototipos. Este modelo aún no debe considerarse como la versión final lista para la competencia global. Los logros y los desafíos encontrados durante este proyecto ayudan a allanar el camino para futuras investigaciones.

Este estudio tiene implicaciones que trascienden los espectros técnicos; Destacan que los efectos sinérgicos entre los entusiastas de la tecnología de investigación pueden despertar el interés por la robótica entre las próximas generaciones. A través de eventos como URUCUP hemos podido formar vínculos con las comunidades locales que nos rodean. Además, compartir nuestra pasión por la robótica entre los miembros de la comunidad ya ha dado frutos al sembrar semillas que potencialmente podrían generar futuros avances en esta fascinante disciplina.

**4 - REFERENCIAS**